 \documentclass[final,3p,times]{elsarticle}

\usepackage{amssymb}

\usepackage{changepage}
\usepackage{lscape}
\usepackage{subcaption}
\usepackage{endnotes}

\journal{arXiv}

\begin{document}

\begin{frontmatter}



\title{The `Paris-end' of town? Urban typology through machine learning.}


\author[melb]{Kerry~A.~Nice\corref{cor1}}
\ead{kerry.nice@unimelb.edu.au}
\author[melb,sunshine]{Jason Thompson}
\author[melb]{Jasper S. Wijnands}
\author[melb]{Gideon D.P.A. Aschwanden}
\author[melb,eng]{Mark Stevenson}

\cortext[cor1]{Principal corresponding author}
\address[melb]{Transport, Health, and Urban Design Hub, Faculty of Architecture, Building, and Planning, University of Melbourne, Victoria 3010, Australia}
\address[eng]{Melbourne School of Engineering; and Melbourne School of Population and Global Health, University of Melbourne, Victoria, Australia.}
\address[sunshine]{Centre for Human Factors and Sociotechnical Systems, University of the Sunshine Coast, Australia.}

\begin{abstract}
The confluence of recent advances in availability of geospatial information, computing power, and artificial intelligence offers new opportunities to understand how and where our cities differ or are alike. Departing from a traditional `top-down' analysis of urban design features, this project analyses millions of images of urban form (consisting of street view, satellite imagery, and street maps) to find shared characteristics. A (novel) neural network-based framework is trained with imagery from the largest 1692 cities in the world and the resulting models are used to compare within-city locations from Melbourne and Sydney to determine the closest connections between these areas and their international comparators. This work demonstrates a new, consistent, and objective method to begin to understand the relationship between cities and their health, transport, and environmental consequences of their design. The results show specific advantages and disadvantages using each type of imagery. Neural networks trained with map imagery will be highly influenced by the mix of roads, public transport, and green and blue space as well as the structure of these elements. The colours of natural and built features stand out as dominant characteristics in satellite imagery. The use of street view imagery will emphasise the features of a human scaled visual geography of streetscapes. Finally, and perhaps most importantly, this research also answers the age-old question, ``Is there really a `Paris-end' to your city?''.

\end{abstract}

\begin{keyword}

machine learning\sep urban typology\sep urban design\sep transport\sep health



\end{keyword}

\end{frontmatter}



\section{Introduction}\label{sec:introduction}

Cities are now home to the majority of the world's population, with trends predicting increasing growth in urbanisation \citep{UNDESA2015,WHO2016,ABS2008}. The top 1700 large cities (with populations exceeding 300,000 residents) contained 2.2 billion people or approximately 31\% of the world's population in 2015 \citep{UN2014}. Continued growth and urbanisation will cause increasing challenges for planners and policy makers to accommodate and provide suitable environments for these populations.

The form a city takes and the way land is allocated can have a detrimental impact on population health and well-being, including car dependency, physical inactivity, and associated illness such as obesity and road trauma \citep{Giles-corti2016,Kleinert2016,Goenka2016,Zapata-Diomedi2017,Heesch2014,Daley2011, Cepeda2016,MingWen2008,Norman2006,Thompson2018b}. Policy-makers and urban/transport planners have an opportunity to reverse this situation by embracing strategies that pro-actively support safe active transport modes as facilitated by urban designs witnessed in some countries around the world. However, understanding the association between urban design features, transport networks, or environmental outcomes remains difficult, especially when underlying data, locations, methods, and demographics upon which statistical models are built vary considerably. As a result, globally consistent comparisons between cities are difficult to achieve. 

Attempts to find quantitative methods, to create city typologies to assess, describe, and classify different types of urban form have been under way for a number of decades. First attempts used broad demographics and functional characteristics to classify different types of cities. Occupational and employment figures were used to determine a city's most important economic activity (including manufacturing, retail, diversified, wholesale, transportation, mining, education, and resorts) \citep{Harris1943}. Other studies used economic activity data to classify cities into broad functional typologies, such as manufacturing, retail, professional services, and financial services \citep{Nelson1955}. \citet{Bruce1971} performed a cluster analysis based on the socio-economic profiles of selected cities as well as a number of census based statistics to group them into clusters. However, in these studies, the resulting typologies are more functional in nature, making the contribution of urban design difficult to examine.

New techniques to define city typologies emerged in the 1980s and 90s with the growing availability of databases of spatial data and increased computing power. Much of this work focused on road infrastructure in cities, and drew from the structural sociology field, in which groups of people were represented as part of a broader network structure. The `space syntax' of \citet{Hillier1996} established a correlation between configurations of urban forms and variations of human interactions within it. 

Other recent remote-sensing based methods depart from the pure network analysis methods to derive urban typologies. Night-time light data has been used to categorise cities into stages of urbanisation and levels of economic activities \citep{Zhang2013}. Urban metrics (road geometry, building dimensions and heights, and vegetation heights) have also been used to classify cities into typologies of differing periods of historical design and urban planning (i.e. 19th Century, 1950s, 1970s, etc.) \citep{Hermosilla2014}.

Building on recent advances in computing power, artificial intelligence, and urban imagery, new approaches have been created to discover unique visual characteristics of cities and how they are used. For example, large numbers of geo-tagged photos have been used to detect patterns of urban usage and public perception of a number of areas' functional and social attributes \citep{Liu2016,Zhou2014a}. Place Pulse, a database of urban imagery using crowd-sourced classifications (including safety, beautiful, and liveliness) has been built to quantify perceptions of urban areas \citep{Dubey2016,Naik2014} and inequality \citep{Salesses2013}. \citet{Doersch2012} used a large number of geo-localised street level images to discover common visual features across a number of cities.

Still, most methods described above require some amount of subjective classification of local input data; the quality and availability of which can vary widely across collection or political districts. To overcome these limitations, we demonstrate a new `bottom-up' approach. We train three neural networks to recognise specific cities using millions of street view, satellite, and digital street map images, and perform a case study using two Australian cities, Melbourne and Sydney. Paris, France is an iconic international city \citep{Anholt2006} with widely recognisable visual elements \citep{Doersch2012}, leading many cities (including Melbourne and Sydney) to claim that they have a `Paris-end' of town \citep{Williams2010}, or are a `Paris on the [insert name of local river]' \citep{Wilden2013} (e.g. `Paris on the Yarra'). This study examines whether Melbourne or Sydney truly claim to have a `Paris-end' and demonstrates a new fundamental methodology to objectively analyse urban areas using big data imagery.

\section{Methods}\label{sec:methods}
\subsection{Neural network}\label{sec:methods1}

The methods applied in this study are based on artificial intelligence, in particular deep neural networks \citep{Bishop1995,Samarasinghe2016,Graupe2013}. Neural network architectures that have proven to be particularly successful at image recognition tasks are convolutional neural networks \citep{Schmidhuber2015}. The model for image recognition used in this study is based on the Inception V2 architecture \citep{Szegedy2015,Ioffe2015}. 

\subsection{Imagery sampling}\label{sec:methods2}

The concept employed in this study was to train a model to correctly recognise individual cities based on examples of different types of urban imagery (street maps, satellite remote sensing, and street view images). The resulting model could then make predictions as to where entirely new images it was presented with were from. Specifically, the assumption was that, if presented with an image of a city that was not Paris but the model `thought' that it was, then the sample city image presumably contained features that were `Paris-like' in nature. 

1692 cities with populations $>$ 300,000 people were initially selected for analysis \citep{UN2014}. Data from Google Maps and Baidu Maps were used to identify urban form for each city in a globally consistent framework. The sampling area for each city was chosen as a circular area aligned to the city's centre, where the radius $r$ (km) of the sampling area was determined based on the population size $p$ according to \citet{Barthelemy2016}: 

\begin{equation}
r = \sqrt{ \frac{28.27}{\pi} \bigg( \frac{p}{300,000}  \bigg)^{0.85} }
\end{equation}

Having identified individual cities, a two-stage sampling approach was applied. As no standardised urban boundaries are available for all the cities evaluated in this study, a methodology had to be developed to define these. Firstly, a sampling area extending 1.5 km from the identified city centroid \citep{UN2014} was set as a baseline. As sample cities' populations increased in size, the sampling area increased by a power of 0.85 to the proportional increase in population size \citep{Barthelemy2016}. Standardising the sampling area in this manner avoided socio-political discrepancies relating to a city's `true' boundary and captured differences in population density and shape between small (e.g., Wellington, New Zealand; Izmit, Turkey) and global mega-cities (e.g., Tokyo, Japan;  Delhi, India). Location sampling areas were adjusted for the earth's curvature \citep{Sinnott1984}. Large water-bodies (e.g., oceans but not coastlines) were removed from the sampling area, as they were not indicative of urban form . 

These procedures result in a population and water body-adjusted circular area centred on the city's central coordinates, capturing the widest extent of each city while minimising the amount of non-urban locations. 

\label{methodsimagery}
\subsection{Imagery sources}

Three neural networks (see Table \ref{tab:neuralnetworks}) were trained using street maps, satellite imagery, and street view imagery from each city. Images were downloaded from each of the following sources, using the appropriate APIs and were randomly sampled for each city and each network. Imagery from Sydney and Melbourne were excluded as they were included in the evaluation dataset. 

\begin{table}[!htbp]
\caption{\bf Neural networks trained and evaluated in this study \label{tab:neuralnetworks}}     
\begin{tabular}{ l l }
 \hline Abbreviation   &  Imagery source \\ \hline
GM & Google Static Maps API, image type of `map'     \\ 
GS & Google Static Maps API, image type of `satellite'      \\
GSV-BSV & Google Street View API/Baidu Street View API     \\ \hline

\end{tabular}
\end{table}

\begin{figure}[!htbp]
    \centering    
\includegraphics[scale=0.5]{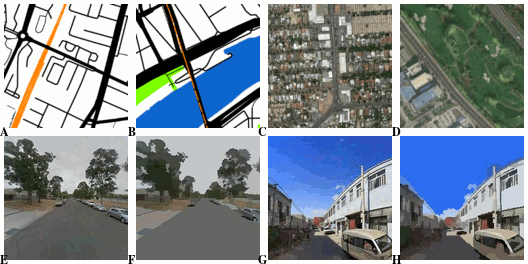}  
\caption{\bf Two sample GM neural network training data images for Paris, France (A, B) \citep{GoogleStatic2017}. Sample GS neural network training data images for Adelaide, Australia (C) and Beijing, China (D) \citep{GoogleStatic2017}. Sample GSV neural network training data image from Sydney, Australia (E) \citep{GoogleMaps2017b} and the processed segmented version (F). Sample BSV neural network training data image from Beijing, China (G) \citep{Baidu2017} and the processed segmented version (H).}    
 \label{fig:maps}  
\end{figure}

The first neural network (referred to as GM) used Google Maps images as training material. Images were sized 256 by 256 pixels using a zoom level of 16 (approximately 400x400m). These were obtained from the selected locations using a custom style defined with the Google Static Maps API \citep{GoogleStatic2017} (see Figure~\ref{fig:maps} for examples of Paris, France). The images provide a high-level abstraction of road (black) and public transport (orange) networks, green space (green), and water bodies (blue). Any remaining space is coded white. Due to mapping inconsistencies in South Korea, all 25 South Korean cities were removed from the dataset, reducing the number of cities to 1665. 1000 training images were used per city (for this neural network as well as the following two), for a total data set of 1,665,000 images in 1665 classifications. 

The second neural network (referred to as GS), used Google Maps satellite imagery obtained through the Google Static Maps API \citep{GoogleStatic2017}. Image type was set to `satellite' using a zoom level of 16 and image size of 256x256. Suitable imagery was not available for two cities, bringing the number of cities to 1688 (also excluding Melbourne and Sydney) and a total data set of 1,688,000 images. Figure~\ref{fig:maps} shows two sample images, one each of Adelaide, Australia and Beijing, China. 

The third neural network (referred to as GSV-BSV) used street view imagery obtained through a combination of Google Street View (GSV) \citep{GoogleMaps2017b} and Baidu Maps Street View (BSV) \citep{Baidu2017}. 1000 images each were sampled for the 1074 cities for which imagery was available (a total of 1,074,000 images) at a 256x256 resolution, a pitch of 0, a field of view of 90 degrees, and a random heading from 0 to 359 degrees. Random headings were used give the imagery the widest range of samples of the urban areas and ensure that the heading itself didn't influence the training (i.e. grid street systems always orientated in the same direction resulting in cities only sampling up and down the centre of streets). Images inside tunnels, indoor locations, dark locations, or otherwise unusable images were removed and replaced by re-sampling.

No street view imagery of China was available through GSV, so BSV was used instead.  In order to minimise the differences between the two data sources and to minimise strong country-specific items (e.g. text on road signs) influencing neural network training, further image processing was performed to segment each image before use in training and evaluation. The Python module \textit{pymeanshift} \citep{Pymeanshift2017} was used to segment each image\footnote{Using a spatial radius of 6, range radius of 4.5, and minimum density of 50.}. Figure~\ref{fig:maps} shows an example of an original GSV image (Sydney, Australia) and its segmented version as well as an original BSV image (Beijing, China) and its segmented version.


Images from Sydney and Melbourne, Australia were excluded from the training data and were instead used for evaluation. This evaluation data was sampled at a 400m grid resolution across the greater metropolitan areas, with 23,027 possible locations for Melbourne and 24,596 for Sydney using the same API methods described above for the training data. Availability of imagery for GSV at these locations was 59.5\% and 91.1\% respectively. The sampled Melbourne area contained a much higher percentage of rural areas without roads (the primary location for GSV imagery) than the sampled Sydney area.

\subsection{Neural network training}\label{sec:methods4}    

The Inception V2 network was used in this study and the three networks (GM, GS, and GSV-BSV) were trained with 256x256 sized imagery. The Inception network was calibrated using supervised learning with the generated dataset to identify the name of the city based on a supplied image. Several pre-processing steps were performed before supplying the image to the neural network. Images were randomly cropped from 256x256x3 to Inception V2's native 224x224x3 resolution. No zooming was applied, the aspect ratio was kept fixed, and colour transformations were not used. All images were normalised to [-1, 1] by subtracting a colour value of 128 from each pixel and multiplying by 1/128. To ensure good mixing, training images were randomly allocated to batches. Validation images (25\% of the 1000 training images for each city were reserved as validation data) were transformed to 224x224x3 using central cropping.

To update weights in the neural network, a loss function was specified to quantify the extent of any current misclassifications, namely the cross entropy calculated on the softmax layer. Model parameters were calibrated by minimising this loss function using Stochastic Gradient Descent with Nesterov momentum of 0.9. Other parameters included a batch size of 64 samples, reducing learning rate starting at 0.9 per batch, batch normalisation, a dropout rate of 0.2 after the final average-pooling operations, and an L2 regularisation weight per sample of 0.0001. Each model was trained until convergence for a total of 150 epochs, using the Microsoft Cognitive Toolkit (CNTK) \citep{Yu2015}. 

\subsection{Neural network inference}\label{sec:methods5}    
Using the three trained models, inferences were performed using the evaluation datasets for Melbourne and Sydney. As Melbourne and Sydney are not present in the training data, the neural network was forced to choose the city with the most similar characteristics for each of the sampled locations. Using these predictions, every location in both cities was determined to be `most like' another world city from the list due to  characteristics contained within the street map, satellite, or street-view image. Note, all neural network classification predictions with a probability lower than 50\% were filtered out of the following results.

\section{Results}\label{sec:results}

Using 25\% of the training data, validation was performed on each model. The models for GM, GS, and GSV-BSV reached a final accuracy of 73.2\% (top 5: 85.0\%), 99.4\% (top 5: 99.97\%), and 43.1\% (top 5: 69.8\%), respectively. These accuracies were calculated at the end of each epoch during the training step, testing the neural network's skill in correctly identifying the correct city out of the nearly 1700 cities (excluding Melbourne and Sydney).

The resulting predictions from model inference of the evaluation data were analysed in various ways. First, the top 20 predicted cities for the evaluation points for each imagery data set were calculated (see Table \ref{tab:melbournesydney20} for GM, GS, and GSV-BSV).

\subsection{Top 20 predicted cities} 

The GM (map view) neural network predictions (Table \ref{tab:melbournesydneyGM}) are dominated by other Australian cities (Brisbane, Canberra, Sunshine Coast, Gold Coast, Newcastle and Lake Macquarie, Perth, and Adelaide) as well as a number of cities from Israel, South Africa, and the United States. Alternative Australian cities make up nearly 20\% of the top 20 predictions for Melbourne and 17\% for Sydney. Melbourne and Sydney also show strong similarities with each other with the neural network considering them similar to the same 12 cities out of the top 20 predictions.

The GS (satellite view) neural network predictions (Table \ref{tab:melbournesydneyGS}) shows wider divergences from other Australian cities and between Melbourne and Sydney themselves, with both often matched to Brazilian cities. Melbourne is matched to Brazil in 11\% of the evaluation locations while Sydney is matched to Brazilian cities in 15\%. Melbourne and Sydney show wider divergences from each other using the GS network in comparison to the GM network, only having 8 of the top 20 predicted cities in common. In diverging predictions, 4.1\% of Melbourne is confused with Wellington, New Zealand while 4.7\% of Sydney is considered similar to Sevastopol, Ukraine. 

The GSV-BSV (street view) neural network predictions (Table \ref{tab:melbournesydneyGSV}) show strong similarities between Melbourne and Sydney. In the Melbourne evaluation, just under 18\% (7 of the top 9 picks) are other Australian cities, while Sydney matched other Australian cities in 20.5\% of the evaluation locations (and were 7 of the top 7 picks) and spread somewhat evenly through these other cities. In addition, 15 of the top 20 predicted cities were shared between Melbourne and Sydney.

\begin{landscape}
\begin{table}[!htbp]
\caption{\bf{Top 20 cities like Melbourne and Sydney. Table shows percentage of evaluated locations predicted to be a certain city for GM, GS, and GSV-BSV neural networks.}}  
\label{tab:melbournesydney20}  
\begin{subtable}{.34\linewidth}
\caption{GM}
\label{tab:melbournesydneyGM}  
\scalebox{0.9}{
\begin{tabular}{ l  l l }
 \hline    
\textbf{Predicted city} &  \textbf{Mel \%}  &  \textbf{Syd \%}\\ \hline
Brisbane, AUS&13.0&	7.9\\
Beer Sheva, ISR&	5.3&	11.0\\
Canberra, AUS	&2.6&	5.7\\
Sunshine Coast, AUS	&2.3&	0.7\\
McAllen, USA&		2.1&		0.3\\
Valpara\'{i}so, CHL&		1.0&	0.7\\
Haifa, ISR&		0.8&1.6\\	
Gold Coast, AUS&		0.8&	0.5\\
Pretoria, ZAF&		0.6&  -      \\
Newcastle, AUS&	0.6&	2.6\\
Toronto, CAN	&0.6&	0.4\\
Auckland, NZL&		0.5&	0.6\\
Perth, AUS&		0.5&  -      \\
Johannesburg, ZAF&		0.4&  -      \\
Philadelphia, USA&		0.4&  -      \\
Jerusalem, ISR&		0.2&	0.8\\
Port Elizabeth, ZAF&		0.3&  -      \\
Washington, D.C., USA&		0.2& -      \\
Southend-On-Sea, GBR&		0.2&  -      \\
Virginia Beach, USA&-&		0.4\\
London, GBR&-&		0.3\\
Barcelona, ESP&-&		0.3\\
Adelaide, AUS&-&		0.3\\
Tasikmalaya, IDN&-&		0.3\\
Douala, CMR&-&		0.2\\
Hardwar, IND&-&		0.2\\
\\
\\
\\
\\
\end{tabular}
}
\end{subtable}%
\begin{subtable}{.34\linewidth}
\caption{GS}
\label{tab:melbournesydneyGS} 
\scalebox{0.9}{
\begin{tabular}{ l   l  l}
 \hline  
\textbf{Predicted city}  & \textbf{Mel \%}  &  \textbf{Syd \%}\\ \hline
Jundia\'{i}, BRA&7.8&	10.6\\
Wellington, NZ&	4.1&		0.09\\
Campinas, BRA&		2.4&		4.0\\
Miami, USA&		2.2&- \\
Adelaide, AUS&		1.7&- \\
Provo-Orem, USA&		1.3&- \\
Macap\'{a}, BRA&		0.6&- \\
Gold Coast, AUS&		0.5&- \\
Rosario, ARG&		0.4&		1.1\\
Qitaihe, CHN&		0.4&- \\
Johannesburg, ZAF&		0.4&		0.05\\
Buenos Aires, ARG&		0.3&- \\
Canberra, AUS&		0.2&- \\
Mendoza, ARG&		0.2&- \\
Juiz De Fora, BRA&		0.2&		0.8\\
Pulandian, CHN&		0.2&- \\
Newcastle, AUS&		0.2&		0.2\\
Palma, ESP&		0.2&		0.08\\
Cape Town, ZAF&		0.2&- \\
Sevastopol, UKR&-&		4.7\\
Memphis, USA&-&		0.6\\
Baaqoobah, IRQ&-&		0.4\\
Belgaum, IND&-&		0.2\\
Qazvin, IRN&-&		0.2\\
Nagasaki, JPN&-	&	0.1\\
New Orleans, USA&-&		0.1\\
Malaga, ESP&-&		0.1\\
Guayaquil, ECU&-&		0.07\\
Khabarovsk, RUS&-&		0.06\\
Chiinu, MDA&-&		0.05\\
\end{tabular}
}
\end{subtable}%
\begin{subtable}{.34\linewidth}
\caption{GSV-BSV}
\label{tab:melbournesydneyGSV} 
\scalebox{0.9}{
\begin{tabular}{ l  l   l}
 \hline      
\textbf{Predicted city} &  \textbf{Mel \%}  &  \textbf{Syd \%}\\ \hline
Sunshine Coast, AUS&		4.0&		5.0\\
Perth, AUS&		4.0&	1.6\\
Adelaide, AUS&		4.0&		1.2\\
Auckland, NZL&		2.7	&0.8\\
Canberra, AUS&		1.9&		3.1\\
Newcastle, AUS	&	1.7&		3.1\\
Gold Coast, AUS&		1.3&	3.3\\
Christchurch, NZL&		0.9&		-\\
Brisbane, AUS&		0.8&	3.2\\
Bonita Springs, USA&		0.7&		0.5\\
Stockton, USA&		0.7&		0.5\\
Curitiba, BRA&		0.7&		0.4\\
Cape Town, ZAF&		0.7&		0.4\\
Concord, USA&		0.6&		0.3\\
East London, ZAF&		0.5&		0.4\\
West Yorkshire, GBR&		0.5&		-\\
Teeside, GBR&		0.4&		-\\
Los Angeles, USA&		0.4&		-\\
Montevideo, URY&		0.3&		-\\
San Diego, USA&		-&		0.9\\
Jacksonville, USA&	-&			0.4\\
Vereeniging, ZAF&	-&	0.4\\
Virginia Beach, USA&		-	&	0.4\\
McAllen, USA&	-	&	0.4\\
\\
\\
\\
\\
\\
\\
\end{tabular}
}
\end{subtable}%
\end{table}
\end{landscape}

To explore the identified differences, cities predicted for an evaluation location were plotted on maps of Melbourne and Sydney, with the colour scheme for the plots determined by the latitude and longitude of the predicted city. This colour scheme is shown in Figure~\ref{fig:colorscheme}.  As such, in the following figures, predicted cities in Australia will show up in shades of yellow, the rest of the Southern Hemisphere in greens, Asia in reds, North America and Europe in blues, and the Middle East in blue/greys.

\begin{figure}[!htbp]
\centering    
\includegraphics[scale=0.25]{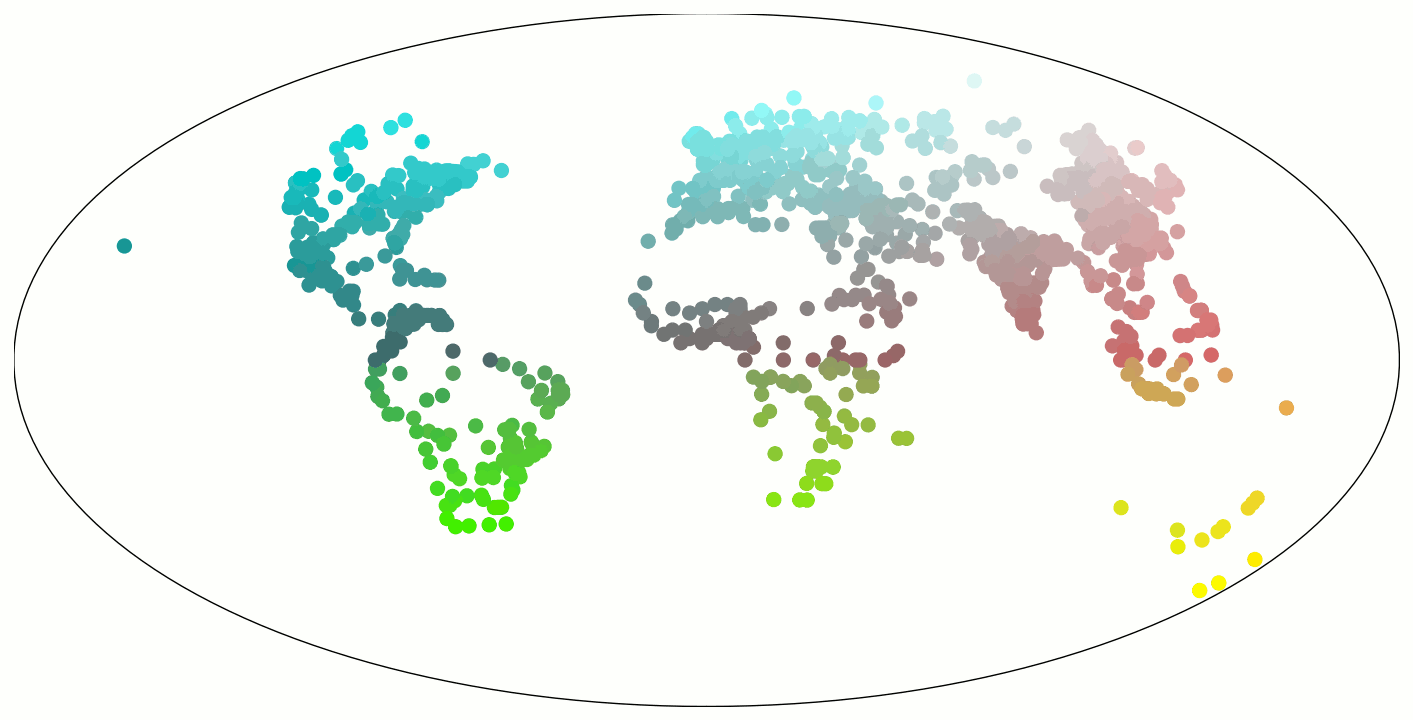} 
\caption{\bf Latitude/longitude based colour scheme for plotting cities-like for Melbourne and Sydney evaluations.}    
 \label{fig:colorscheme}  
\end{figure} 

\subsection{Melbourne evaluation} 

\begin{figure}[!htbp]
\centering   
\includegraphics[scale=0.10]{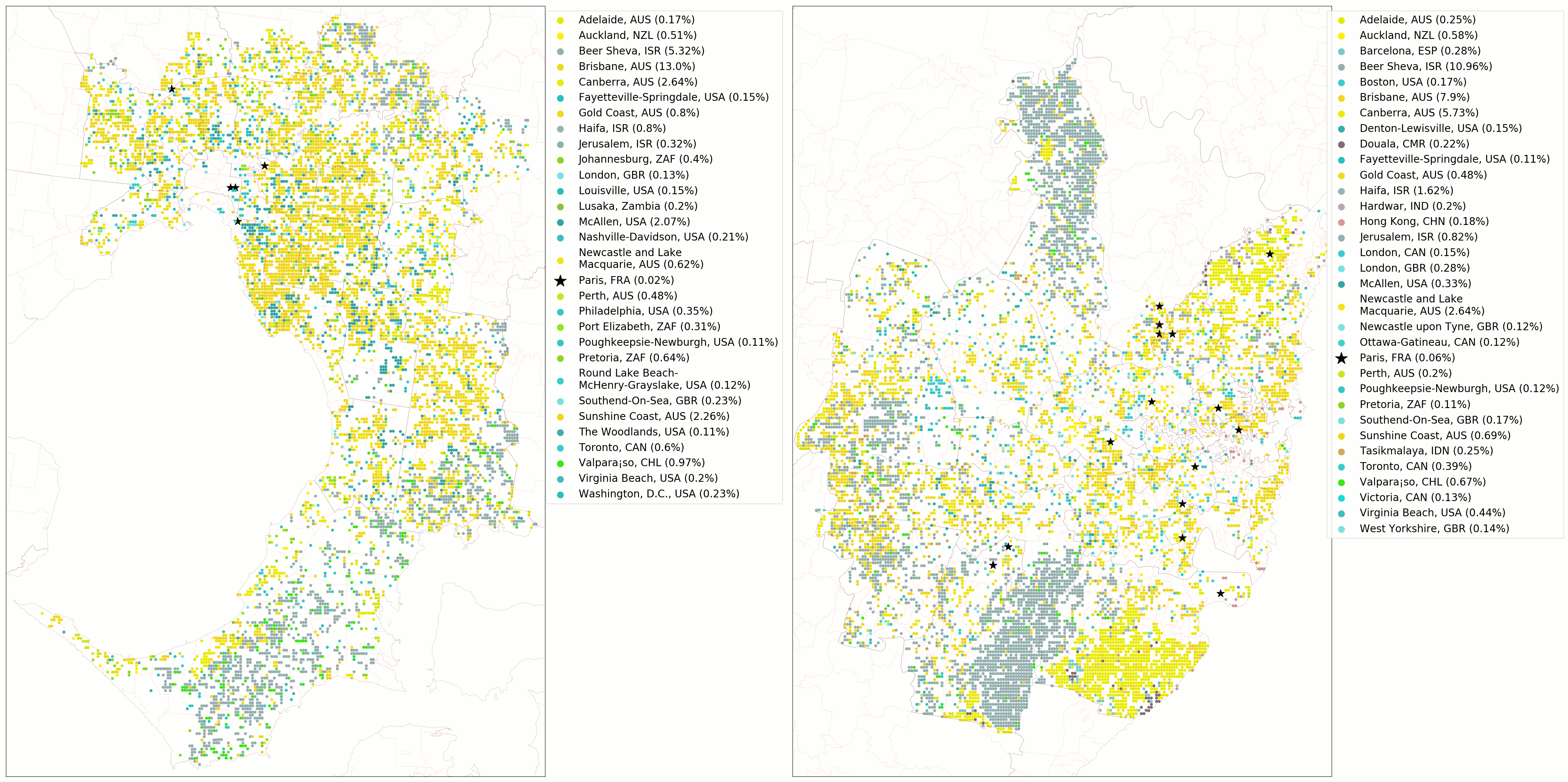}  
\caption{  \bf Predicted similar cities using the GM neural network (filtering out probabilities lower than 50\%). Top predicted cities plotted (using Figure~\ref{fig:colorscheme} colour scheme) for Melbourne evaluation locations (left) and Sydney (right). Predicted Paris locations marked with black stars.}    
 \label{fig:melmaps}  
\end{figure}

Figure~\ref{fig:melmaps} shows the top predicted cities ($>$ 0.1\%) plotted against the Melbourne evaluation locations for the GM neural network. Further, `Paris-like' evaluation locations within Melbourne and Sydney are highlighted with black stars (22 in total, but 5 with probabilities greater than 50\%). As can be seen, Australian cities (in yellow) show strong groupings in the inner and outer suburbs while the central business district (CBD) region shows no single strong grouping of regions or specific cities. In Melbourne's far outer suburbs and rural areas, a wide mix of North and South American, South African, European, and Mid-Eastern cities (in greens blues and greys) with small localised clusters of each can be seen. In the CBD, a few locations are predicted as Paris, and are mostly associated with Docklands or parklands.

\begin{figure}[!htbp]
\centering   
\includegraphics[scale=0.10]{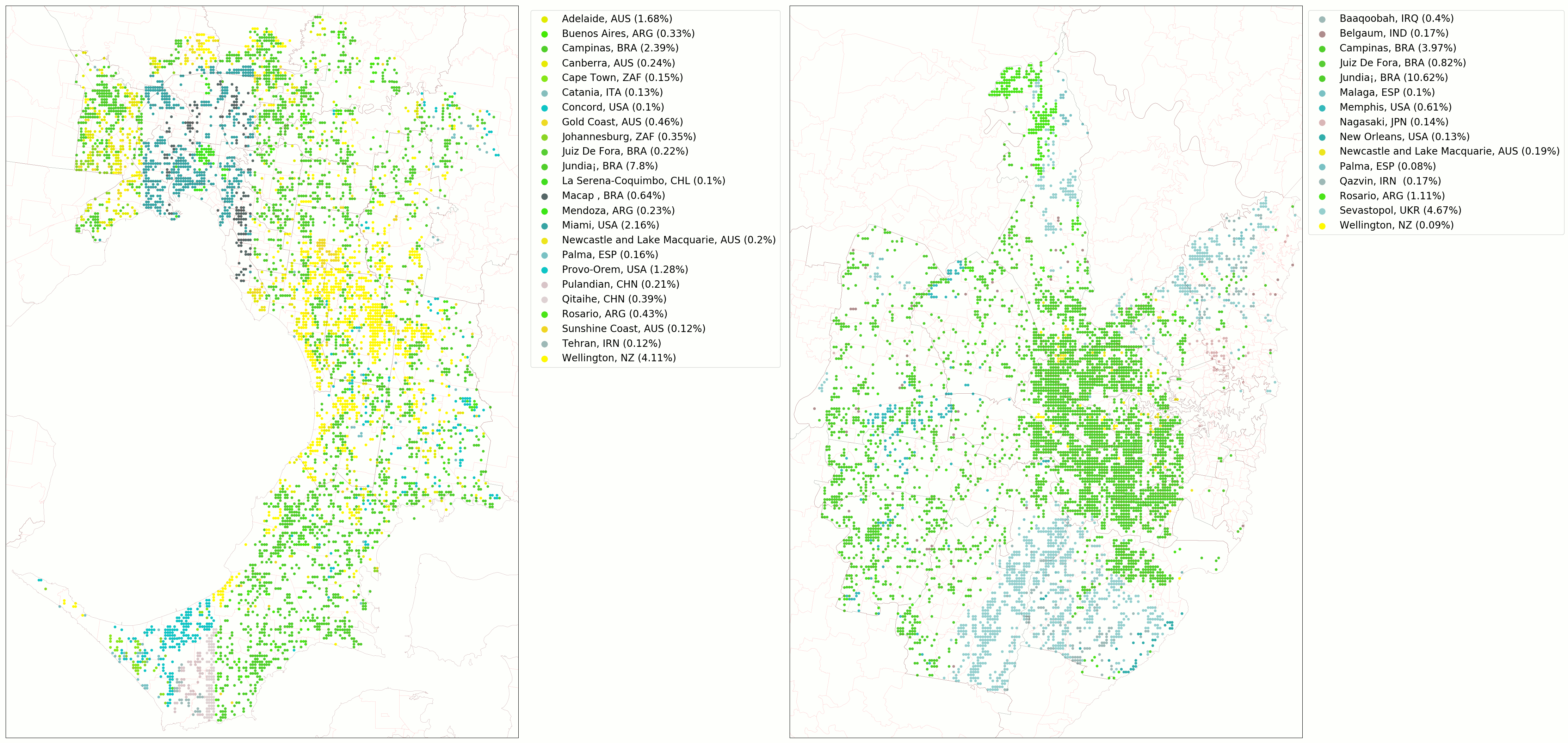}  
\caption{ \bf Predicted similar cities using the GS neural network (filtering out probabilities lower than 50\%). Top predicted cities plotted (using Figure~\ref{fig:colorscheme} colour scheme) for Melbourne evaluation locations (left) and Sydney (right). Predicted Paris locations marked with black stars.}    
 \label{fig:melsat}  
\end{figure}

Figure~\ref{fig:melsat} shows the top predicted cities ($>$ 0.1\%) plotted against the Melbourne evaluation locations for the GS neural network with `Paris-like' locations again highlighted with a black star (1 location, but 0 locations above 50\% probability). Other Australian cities (yellows) show a strong grouping in the inner and outer suburbs while the CBD region shows no single strong grouping of regions or specific cities but with a range of predictions including Miami, United States (blues) and Mendoza, Argentina (greens). In Melbourne's far outer suburbs and rural areas, a wide mix is seen of North and South American (USA, Brazil, and Argentina), South African, European (Italy and Spain), and Mid-Eastern (Iran and Turkey) cities with small localised clusters of each. Only a single predictions of Paris, France was made by the GS neural network for any evaluation location in Melbourne (but not above a 50\% probability).

\begin{figure}[!htbp]
\centering   
\includegraphics[scale=0.10]{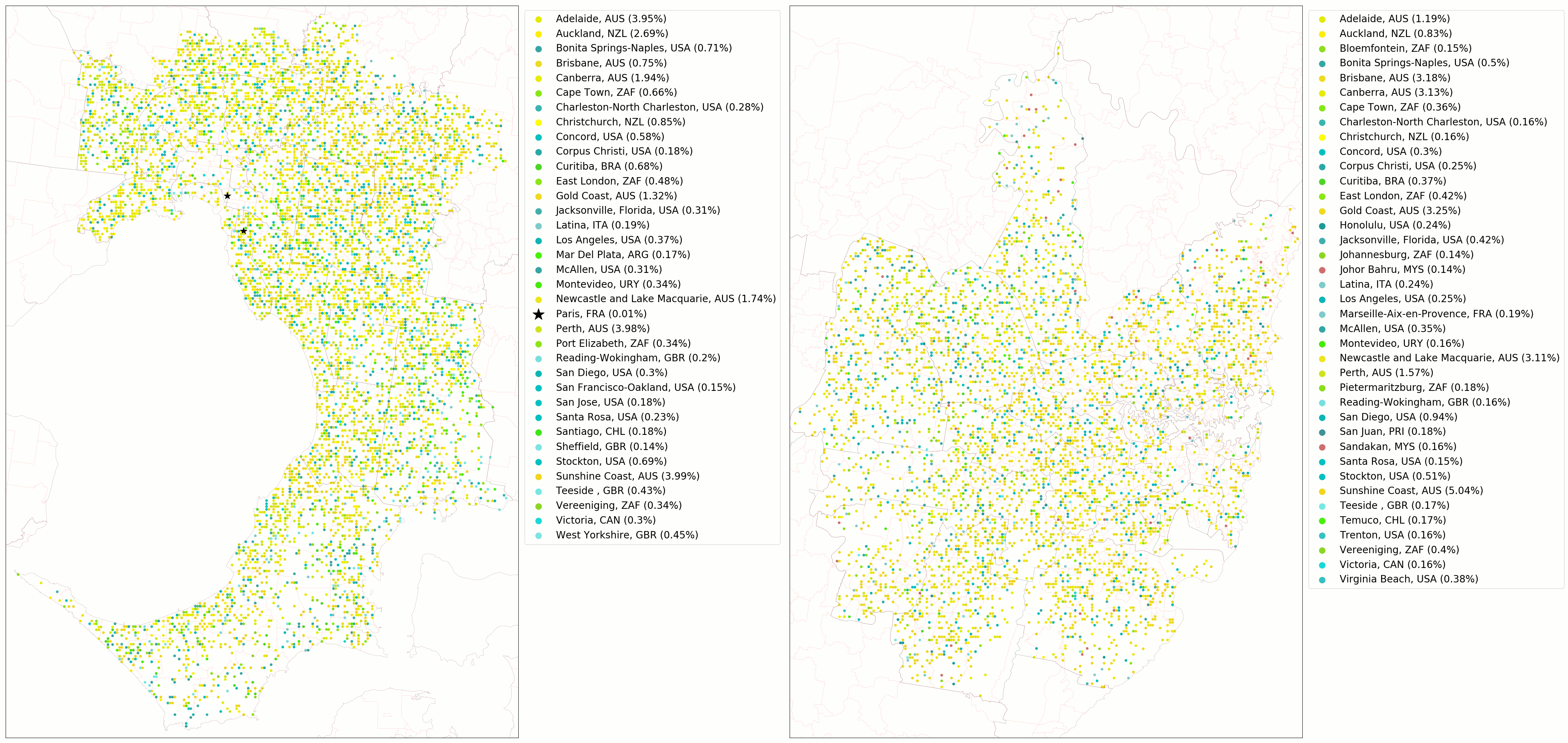} 
\caption{ \bf Predicted similar cities using the GSV-BSV neural network (filtering out probabilities lower than 50\%). Top predicted cities plotted (using Figure~\ref{fig:colorscheme} colour scheme) for Melbourne evaluation locations (left) and Sydney (right). Predicted Paris locations marked with black stars.}    
 \label{fig:melstreet}  
\end{figure}

Figure~\ref{fig:melstreet} shows the top predicted cities ($>$ 0.1\%) plotted against the Melbourne evaluation locations for the GSV-BSV neural network. `Paris-like' locations are predicted in 13 locations (but only 2 with a probability over 50\%). The overall predictions are dominated by other Australian cities (yellows) scattered widely throughout the entire greater Melbourne area. The remaining evaluation locations show no strong groupings of any predicted countries or cities. Common predictions include cities from South Africa (greens), New Zealand (yellows), the United States, and European countries (blues). The CBD again shows a wide scattering of predictions with no dominant single city or country.

\subsection{Sydney evaluation} 

Figure~\ref{fig:melmaps} shows the top predicted cities ($>$ 0.1\%) plotted against the Sydney evaluation locations for the GM neural network. `Paris-like' areas are predicted in 54 locations (but only 15 above 50\% probability).  Alternative Australian cities (yellows) appear in the western and south eastern suburbs, while Mid Eastern cities (greys) tend to appear in northern and southern suburbs. The CBD and central parts of the city show less single city or regional groupings but with stronger highly localised clusters of each. Some cities commonly represented in the CBD include waterfront cities such as Hong Kong, London, Toulon, and Kaohsiung. 

Figure~\ref{fig:melsat} shows the top predicted cities ($>$ 0.1\%) plotted against the Sydney evaluation locations for the GS neural network. The overall predictions are dominated by cities in Brazil and other South American locations (greens) in the north, west, and central regions, and Ukraine (blues) in the south. Other Australian cities are only predicted in a few locations around the city. In the CBD, predictions continue to be dominated by Brazilian cities with some more scattered predictions of cities from Japan, Haiti, and Mexico. No predictions of Paris, France were made by the GS neural network for any evaluation location in Sydney.

Figure~\ref{fig:melstreet} shows the top predicted cities ($>$ 0.1\%), plotted against the Sydney evaluation locations for the GSV-BSV neural network. Six `Paris-like' locations were predicted (but zero with probabilities greater than 50\%). Results are very similar to the Melbourne evaluation. Again, the overall predictions are dominated by other Australian cities scattered widely throughout the entire greater Sydney area. The remaining predicted results show no strong groupings of any predicted countries or cities but some of the common predictions include cities from the United States, New Zealand, South Africa, and a number of European countries. The CBD shows a similar scattering of predictions with no single city or country dominating. A summary of the predicted `Paris-like' locations across all three neural networks for each city are presented in Table \ref{tab:melbournesydneyparis}.

\begin{table}[!htbp]
\caption{\bf How much are Melbourne and Sydney like Paris? Table shows neural network results unfiltered and filtered (where predictions with probabilities less than 50\% are removed).  \label{tab:melbournesydneyparis}}     
\begin{tabular}{ l  l l l  l}
 \hline    &  \multicolumn{2}{c}{\textbf{Melbourne}} & \multicolumn{2}{c}{\textbf{Sydney}}  \\  
\textbf{Neural network} & \textbf{Matches} & \textbf{\% matching}  & \textbf{Matches} & \textbf{\% matching}\\ \hline
GM & 22 & 0.1 & 54 & 0.22 \\ 
GM (filtered $<$ 50\%) & 5 & 0.02 & 15 & 0.06 \\ 
GS & 1 & 0.00004 & 0 & 0 \\ 
GS (filtered $<$ 50\%)& 0 & 0 & 0 & 0 \\ 
GSV-BSV & 13 & 0.06 & 6 & 0.04 \\ 
GSV-BSV (filtered $<$ 50\%)& 2 & 0.01 & 0 & 0\\
\hline
\end{tabular}
\end{table}

\section{Discussion}\label{sec:discussion}
This study sought to answer ``is there a `Paris-end' of Melbourne or Sydney"? As the results show, we can conclusively state that neither Melbourne or Sydney have a strong case to claim that they are like Paris or have an extensive `Paris-end' of town. Using three different trained neural networks and three different sources of imagery, very few locations in Melbourne or Sydney are confused with Paris by the neural networks. However, the process of answering this question served as a demonstration of how the combination of urban imagery and neural networks can be used in constructing urban typologies.

\begin{figure}[!htbp]
\centering   
\includegraphics[scale=0.10]{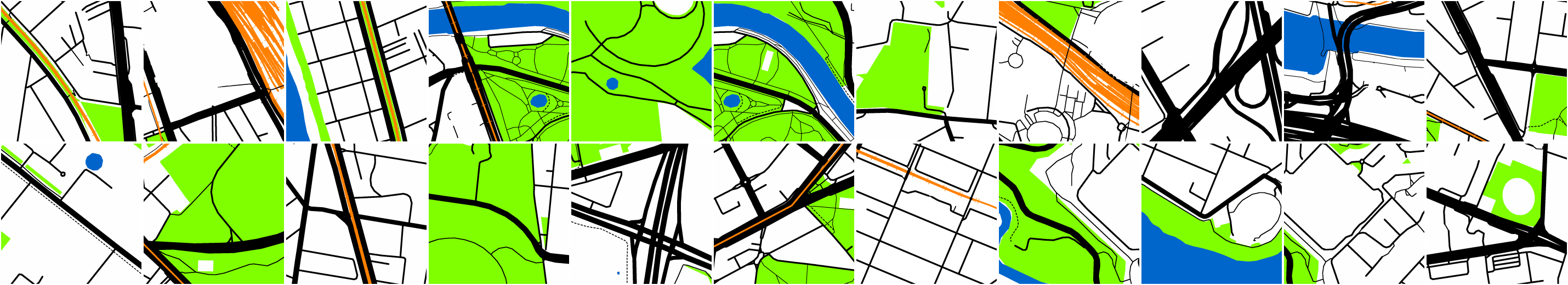}   
\caption{\bf Gallery of `Paris-like` locations in Melbourne using the GM neural network.}    
 \label{fig:gm_mel_gallery} 
\end{figure}

\begin{figure}[!htbp]
\centering   
\includegraphics[scale=0.11]{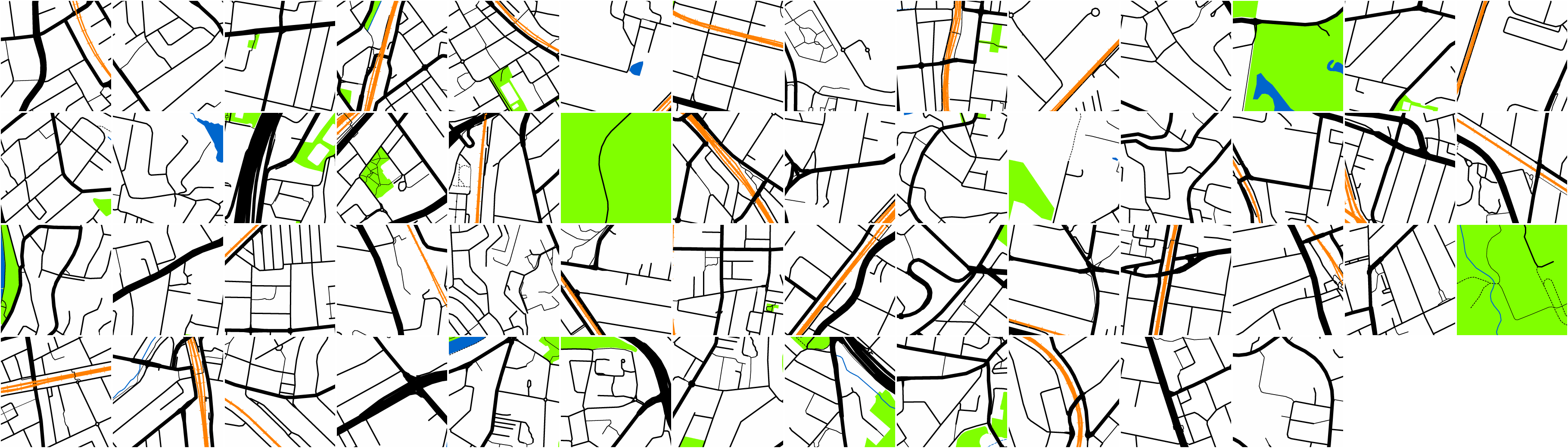}   
\caption{\bf Gallery of `Paris-like` locations in Sydney using the GM neural network.}    
 \label{fig:gm_syd_gallery}  
\end{figure}

In looking at the few locations that are deemed to be `Paris-like', there are a number of common characteristics that stand out. A gallery of all of the images for Melbourne and Sydney that the GM neural network found were similar to Paris are presented in Figures~\ref{fig:gm_mel_gallery} and \ref{fig:gm_syd_gallery}. There are a number of common elements in these images. Many show large parklands (in green) embedded in the cities. Orange lines of public transport (rail and tram) are also prominent as well as large water bodies (in blue). Large arterial and trunk roads run nearby smaller (often curving) local roads, however these local roads tend to still be larger and do not reach the small intricate layouts of some Asian cities. The GM neural network is making predictions based on mapping imagery, capturing characteristics such as the mix and detail of public transport, green space, water bodies, and the road network structure. This includes whether the roads are grid-like, the mix of arterial vs. neighbourhood roads, and their integration with the rest of the urban form. Seven Australian cities were included in the training data (Perth, Brisbane, Sunshine Coast, Gold Coast, Newcastle and Lake Macquarie, Canberra, and Adelaide) and likely share many common planning and design standards with Sydney and Melbourne, influencing the neural network's predictions. 

Using the GS neural network, none of the evaluated locations for Sydney and only one location for Melbourne were predicted to be `Paris-like'. From an overhead remote sensing point of view, there is therefore nothing about either Melbourne or Sydney that shares similar visual characteristics with Paris, or at least there are many other cities that are more similar to Paris than Melbourne and Sydney. The GS network is more strongly influenced by larger natural and topographical features (features visible through satellite imagery) than the GM network. Outside of the immediate city centres, both Melbourne and Sydney are highly vegetated, with large percentages of the built-form concealed under tree canopies and having to conform to topography. The colours of the vegetation and soils as well as how the urban form is mixed into the canopies, hillsides, waterways and oceans are highly influential. Melbourne is built around a bay and around a north-south spine of hills while Sydney is built around the open ocean and ocean waterways as well as hilly terrain throughout the metro area. Some potential limitations in the dataset can be seen in Figure~\ref{fig:melsat}. A strong north-south gradient through the plot of the Melbourne predictions suggest that the neural network detected some artefacts of the satellite imagery gathering process, such as different acquisition times of the imagery, that were not apparent to human observation. 

\begin{figure}[!htbp]
\centering    
\includegraphics[scale=0.20]{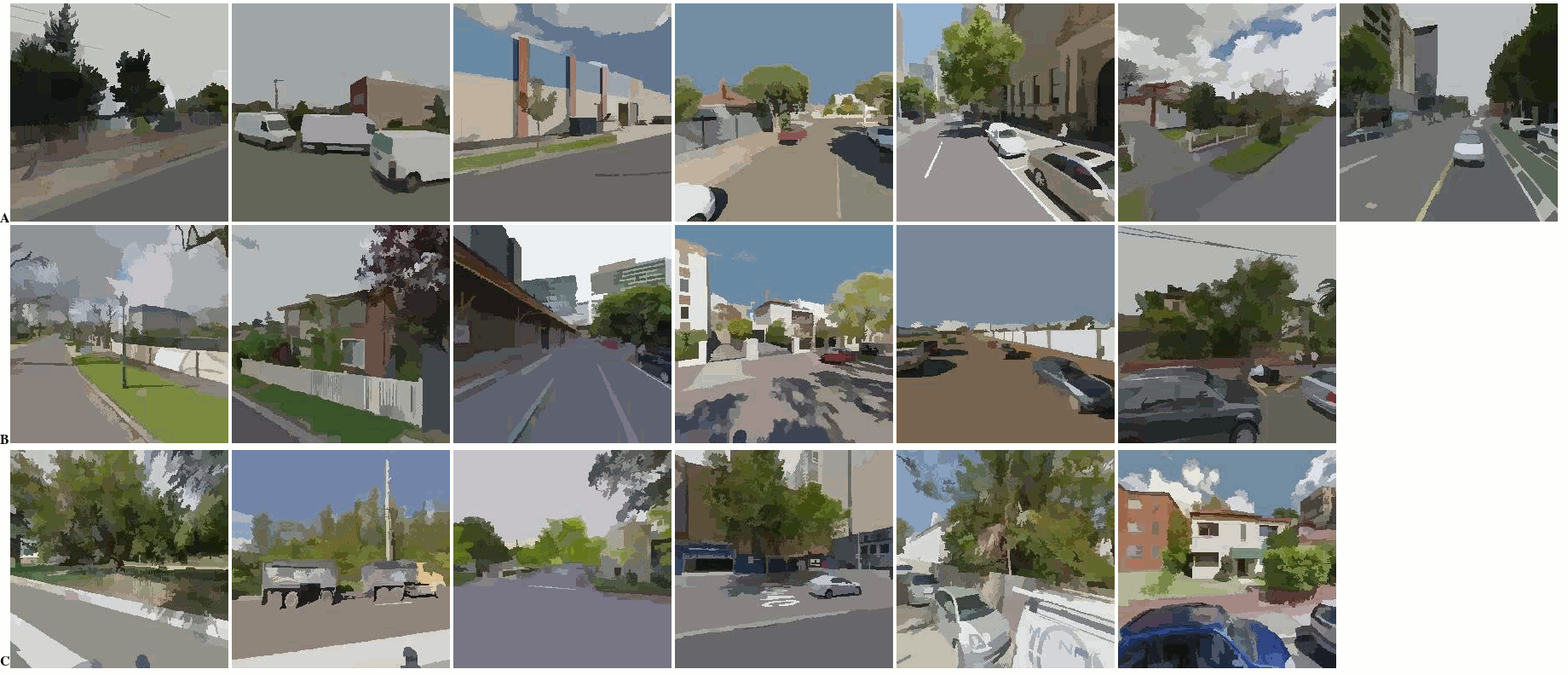}  
\caption{\bf Gallery of `Paris-like` locations in Melbourne (A, B) and Sydney (C) using the GSV-BSV neural network.}    
 \label{fig:gsv_mel_gallery}  
\end{figure}

Finally, as the GSV-BSV neural network only picked Paris (at over a 50\% probability) for 0.01\% of the evaluated locations for Melbourne and 0\% for Sydney, we can be confident that from a visual street-level view, there is almost nothing about either Melbourne or Sydney that is visually similar to Paris using this type of imagery. Of the images for Melbourne, only 2 (out of 13) were picked with a probability of over 50\% (and 0 out of 6 for Sydney). With the GSV-BSV network (galleries of `Paris-like' images for Melbourne and Sydney are shown in Figure~\ref{fig:gsv_mel_gallery}), smaller details of the cities will influence predictions. At this level of imagery, many of the natural features influential in the GS network (e.g., types and colours of vegetation or soil) will be important but smaller details will also weigh in, such as building architecture, the width (or absence) of nature strips or sidewalks, and an overall density of streetscape features. Other influential characteristics are features that are in the urban areas but are not part of the permanent built form. For example, white vans feature in a number of images in the galleries of Paris-like predictions. At this level of imagery, the neural network will be potentially influenced as much by how the urban form is being used as the form itself. This shows the importance of taking steps in some circumstances to construct abstract features from the source images (e.g. road networks/green space for GM or image segmentation for GSV-BSV). Even with these measures, some caution should be taken with this type of imagery. The rather low accuracy rate for GSV-BSV (43.1\%, top 5: 69.8\%) indicates that larger training datasets or perhaps fewer classification classes are needed with this type of complex imagery.

Using the GM neural network approach, urban form can be evaluated. Map characteristics that are influential in grouping cities with a particular typology include extents and types of public transportation, urban green space, road network structure, water body inclusion and integration, amounts of informal unplanned open space, and density and topology influences on city structure. Some of the features included in the GM imagery that made cities `Paris-like' were a higher density of trains and trams, large broad sections of urban green space, and an integration of urban green space and waterways. Of course, while Paris was selected as the comparison city of choice, the technique makes it possible to typify the characteristics of any global city where similar imagery is available.

\begin{figure}[!htbp]
\centering    
\includegraphics[scale=0.30]{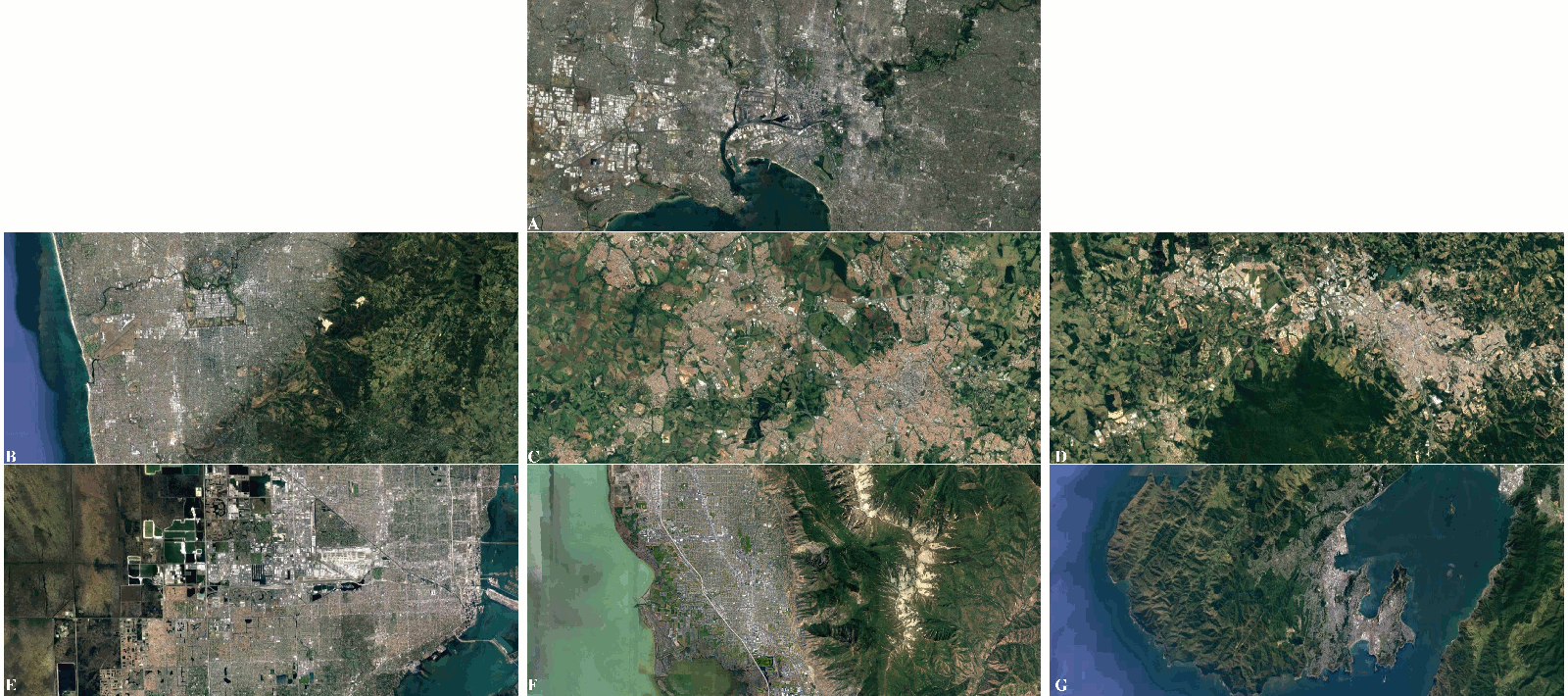} 
 \caption{\bf Satellite imagery of Melbourne, Australia (A), Adelaide, Australia (B), Campinas, Brazil (C), Jundia\'{i}, Brazil (D), Miami, USA (E), Provo, USA (F), and Wellington, NZ (G) \citep{GoogleStatic2017}.}    
 \label{fig:satimages}  
\end{figure}

Using satellite imagery, natural features and the colour characteristics of rooftops, streets, soil, and vegetation feature predominantly in classifying locations within a particular typology. In Figure~\ref{fig:satimages}a, satellite imagery of Melbourne shows a number of colour and terrain similarities with the GS top 6 predictions, namely Adelaide, Australia; Campinas, Brazil; Jundia\'{i}, Brazil; Miami, USA; Provo, USA; and Wellington, NZ (all shown in Figure~\ref{fig:satimages}). This perhaps shows that natural characteristics are more influential to what the GS neural networks considers make cities similar than the characteristics of built urban form highlighted by the GM model.

Finally, in examining the results from the GSV-BSV neural network, this micro-scaled level of imagery would arguably capture the visual geography of the streetscape, what most people would say `this is what makes Paris look like Paris'. But \citet{Doersch2012} found in trying to answer the same question, overall this answer is not based on a small number of famous iconic landmarks (i.e. the Eiffel Tower, the Louvre, etc.), but on an array of widespread, smaller features. These features include elements such as cast-iron railings on balconies, grid-like balcony arrangements, distinctive street signs, streetlamps on pedestals, window balustrades, Parisian doorways, six story Haussmann apartment buildings, and vegetation differences \citep{Li2015}. Of all these micro-scaled visual elements, neither Melbourne nor Sydney contain enough to truly have a `Paris-like' district.

Also, as found in this study, the characteristics that make up a city on a visual street view level are a complex mix. This not only includes bigger structural details, buildings, roads, cars, vegetation, and street furniture, but also smaller less apparent details such as colours, weather conditions, road markings, and thousands of other small details. The complexity of this imagery and the subsequent low accuracy of the neural networks in identifying individual cities using it indicates that further steps are needed to use this type of imagery successfully. These steps can include training using a smaller pool of cities, a smaller set of classifications to allow focus on more subtle differences.

This project was intended to demonstrate the ability of this new methodology to compare and cluster entire cities based on the summation of smaller localised details of urban form. As such, the imagery sampling collected imagery from the entire wider city and not restricted to the perhaps more distinctive city centres. The results reflect that focus and shows one of the strengths of this technique, allowing comparisons between entire cities and allow linkages to datasets (health, transportation, etc.) that exist at city levels. To compare smaller regional portions of cities, the sampling and training methodology merely needs to reduce the sampling radius.

Future work is planned to vary these techniques and further evolve the insights gained. Inner-city comparisons will sample imagery from within cities and help answer questions such as does (wider) Paris look like (the iconic districts of) Paris? Or removing all the other Australian cities from the training data will allow comparisons to be made on a strictly international basis. Cross-comparisons can also assess similarities between individual cities under different contexts (e.g. varying which other cities are included in the pool of comparison cities). 

\section{Conclusion}\label{sec:conclusion}

We have conclusively answered the question, does Melbourne or Sydney have a `Paris-end' of town with a definitive `no'. Our three trained neural networks concluded that at best, Sydney could be considered only 0.06\% `Paris-like' while Melbourne can only boast 0.02\%.   

Despite these potentially disappointing results, this analysis reveals a number of exciting possibilities for using neural networks to analyse urban form. Using this method, any city in the world can now also answer this question for themselves with easily obtained and globally consistent imagery. This methodology can be used to look at many different aspects of cities and understand what elements of their urban design leads them to work in different ways.


\section*{Acknowledgements}
This project was made possible thanks to computer hardware purchased by the Transportation, Health, and Urban Design (THUD) Hub at the University of Melbourne. M.S. was supported by a National Health and Medical Research Council (Australia) Fellowship.

\section*{Data Availability}\label{sec:methods6} 
The datasets, including raw data and trained neural network models, are available on request to the corresponding author.

\section*{Author contributions statement}

K.N. designed and performed the experiment, analysed the results, and wrote the manuscript. J.T. conceived the experiment and contributed to the manuscript. J.S. designed the neural networks and contributed to the manuscript. G.A. contributed to the manuscript. M.S. reviewed the experiment and results. All authors reviewed the manuscript.

\section{References}\label{ref}

\bibliography{ParisEnd-arXivPreprint}
\bibliographystyle{elsarticle-harv} 

\end{document}